\documentclass[runningheads]{llncs}

\usepackage{eccv}

\usepackage{eccvabbrv}

\usepackage{graphicx}
\usepackage{booktabs}
\usepackage{amssymb}
\usepackage{bbding}
\usepackage{subcaption}

\usepackage{multirow}
\usepackage[accsupp]{axessibility}  

\usepackage{hyperref}

\usepackage{orcidlink}

\newcommand{\methodname}{ReGround3D}
\newcommand{\taskname}{3D reasoning grounding}
\newcommand{\mechanismname}{Chain-of-Grounding}
\newcommand{\tabincell}[2]{\begin{tabular}{@{}#1@{}}#2\end{tabular}}

\begin{document}

\title{ScanReason: Empowering 3D Visual Grounding with Reasoning Capabilities} 

\titlerunning{ScanReason}

\author{Chenming Zhu\inst{1,2} \and
Tai Wang \inst{2} \and Wenwei Zhang \inst{2} \and Kai Chen \inst{2} \and
Xihui Liu\inst{1\dag}}

\authorrunning{C.~Zhu et al.}

\institute{The University of Hong Kong \and
Shanghai AI Laboratory\\
\email{chaimzhu@connect.hku.hk} \\ 
\url{https://zcmax.github.io/projects/ScanReason}}

\maketitle

\renewcommand{\thefootnote}{\dag}
\footnotetext[1]{Corresponding author: xihuiliu@eee.hku.hk}

\begin{abstract}

Although great progress has been made in 3D visual grounding, current models still rely on explicit textual descriptions for grounding and lack the ability to reason human intentions from implicit instructions.
We propose a new task called \taskname{} and introduce a new benchmark ScanReason which provides over 10K question-answer-location pairs from five reasoning types that require the synerization of reasoning and grounding. 
We further design our approach, \methodname{}, composed of the visual-centric reasoning module empowered by Multi-modal Large Language Model (MLLM) and the 3D grounding module to obtain accurate object locations by looking back to the enhanced geometry and fine-grained details from the 3D scenes. A chain-of-grounding mechanism is proposed to further boost the performance with interleaved reasoning and grounding steps during inference. Extensive experiments on the proposed benchmark validate the effectiveness of our proposed approach.

\end{abstract}

\section{Introduction}
\label{sec:intro}

\begin{figure*}[t]
  \centering
   \includegraphics[width=1\linewidth]{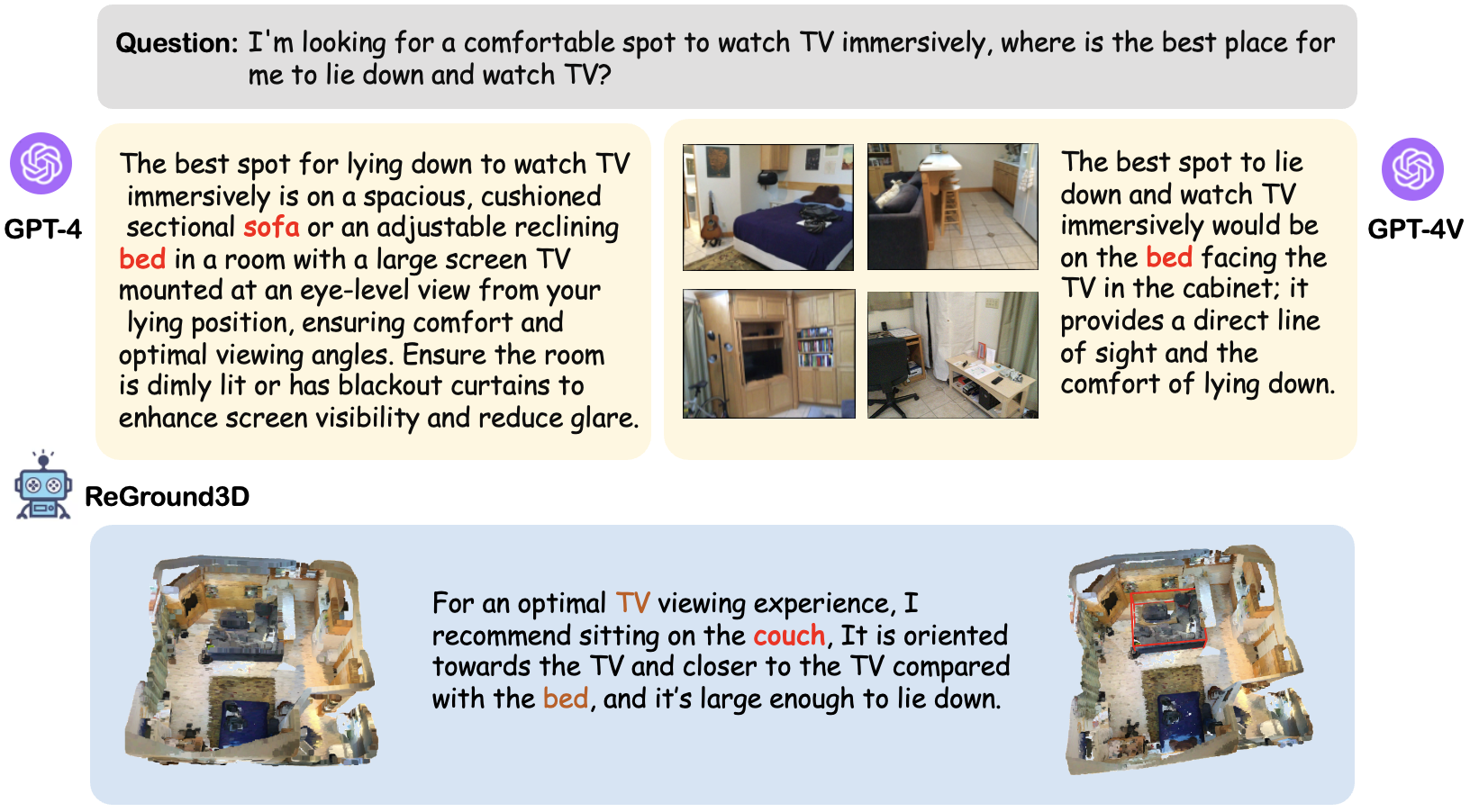}
   \caption{For an embodied agent, they not only need to be able to understand the 3D environment and complex human instructions but also localize the target objects for interaction and navigation. Although GPT-4 (GPT-4V) have strong text (multi-modal) reasoning abilities, they lack the ability to directly perceive the 3D scene, understand the 3D spatial relationships, and output corresponding target object locations. Instead, our proposed method ReGround3D has the 3D perception, reasoning, and grounding capabilities in the real-world 3D environment.} 
   \label{fig:teaser}
\end{figure*}

Understanding and reasoning in the 3D visual world is critical for applications such as robotics and AR, where embodied agents are expected to understand the 3D layout and predict the 3D locations of objects based on human instructions. The example in~\cref{fig:teaser} demonstrates a scenario where the question can only be solved with a comprehensive understanding of the 3D scene and joint reasoning of the question and the 3D environment. However, current 3D visual grounding models~\cite{jain2022bottom, wu2022eda, zhu2023object2scene, yang2021sat} trained on~\cite{chen2020scanrefer,achlioptas2020referit3d} localize objects based on explicit descriptions of the object category, attribute, and 3D spatial relationships, and lack the ability to reason the user intentions and predict object locations with implicit human instructions such as ``I am thirsty, can I have something to drink?''.

To bridge the aforementioned gap and to push the boundaries of what embodied agents can understand and how they can interact with the 3D world, we propose a new task of \taskname{} and introduce a new benchmark named \textbf{ScanReason}. The task requires the model to conduct joint reasoning on the question and the 3D environment before predicting the 3D locations of target objects. We define five categories of 3D reasoning: spatial reasoning, functional reasoning, logical reasoning, emotional reasoning, and safety reasoning. The first two categories focus on the fundamental understanding of the 3D physical world, while the last three categories are built upon fundamental abilities to address user-centric real-world challenges. The benchmark comprises more than 10K question-answer-3D bounding box pairs from 2K scenes belonging to the five reasoning types mentioned above. The GPT-4-assisted data annotation process largely increases the efficiency of curating such a dataset.

We propose ReGround3D as an initial attempt to the new task of 3D Reasoning Grounding. 
Intuitively, for 3D grounding with implicit instructions, we need to first conduct reasoning on the language instructions and the coarse visual environment. Then with the idea of which object we want to find in mind, we look back to the 3D scene and ground the target object. For complex instructions, we may need to alternate the reasoning and look-back process for multiple iterations.
Inspired by this intuition, our framework is composed of a visual-centric reasoning module and 3D grounding with geometry-enhanced look-back module, with a Chain-of-Grounding mechanism during inference to alternately conduct reasoning and grounding for multiple rounds.
Specifically, the visual-centric reasoning module conducts joint reasoning of the 3D scene and instructions with an MLLM. This module predicts a special token representing the semantic and location information of the target object, which is used for the grounding module. The 3D grounding module uses the output token embedding from the previous reasoning module to locate the target object by looking back at the fine-grained 3D scene representation. Unlike previous MLLMs attempting to directly predict bounding box coordinates, our look-back mechanism enables the model to capture more comprehensive 3D geometry and fine-grained object details for accurate 3D grounding. The Chain-of-Grounding mechanism is proposed to synergize reasoning and grounding, which allows multiple rounds alternating between reasoning and grounding during inference.

In summary, our contributions are threefold: 1) We propose a new task of \taskname{} which requires the model to synergize reasoning and grounding. We further introduce a new benchmark ScanReason, which comprises five reasoning types (spatial reasoning, functional reasoning, logical reasoning, emotional reasoning, and safety reasoning) for the task of 3D reasoning grounding in 3D scenes.
2) We design a new framework \methodname{} with a visual-centric reasoning module and a 3D grounding module with geometry-enhanced look-back. We further introduce a Chain-of-Grounding mechanism to boost the \taskname{} ability with a chain of interleaved reasoning and grounding steps.
3) Extensive experiments demonstrate the effectiveness of the our \methodname{} on the ScanReason benchmark for \taskname{}.

\section{Related Work}
\label{sec:related_work}

\subsubsection{3D Vision and Language Learning}
3D Vision-language learning (3D-VL) is garnering increasing attention, with many 3D-VL tasks focusing on how to connect the 3D world with natural language. Among them, 3D Question Answering (3D QA)~\cite{azuma2022scanqa, ye20223d} aims to enable models to provide text answers based on natural language questions. Situation Question Answering in 3D Scenes (SQA3D)~\cite{ma2022sqa3d} requires an agent to first understand its location based on a text description, then provide a reasonable answer based on the surrounding environment, which can be seen as an extension of 3D QA in the embodied AI area. 3D Visual Grounding~\cite{achlioptas2020referit3d, chen2020scanrefer, luo20223dsps, wu2022eda, baroni2020linguistic, roh2022languagerefer} demands that models identify and locate target objects in a 3D scene based on given descriptions, outputting the objects' coordinates and 3D bounding boxes. These descriptions usually explicitly rely on the objects' attributes and their spatial relationships. 3D Dense Captioning~\cite{chen2023vote2cap-detr, chen2021scan2cap, zhong2022contextual3DdenseCap, yuan2022x-trans2cap, wang2022spacap3d, jiao2022more, chen2021d3net, cai20223djcg} requires models to output a series of object coordinates and corresponding scene-based descriptions based on a given scene. Different from these 3D-VL tasks, the questions in \taskname{} could be more implicit and complex.

\subsubsection{3D Visual Grounding} 
The task of 3D visual grounding is aimed at localizing the objects that are explicitly referred to by free-form guided language expressions in the 3D scene. Inspired by the success of transformers in natural language processing, recent 3D visual grounding approaches~\cite{wu2022eda, jain2022bottom, zhu2023object2scene, zhu20233d-vista, chen2022vil3dref, chen2023unit3d} have started to adopt transformer~\cite{transformer} architectures for handling the complex relationships between language descriptions and 3D visual data. These methods leverage the self-attention mechanism of transformers to dynamically weigh the importance of different parts of the input data, facilitating a more effective grounding of textual descriptions in the 3D environment. Recent method~\cite{bakr2023cot3dref} proposes a Chain-of-Thoughts module that predicts a chain of anchor objects that are subsequently utilized to localize the final target object. Compared with 3D visual grounding, our proposed \taskname{} requires the model to reason the complex question, ground target objects, and give the explanation at the same time.

\subsubsection{Multi-modal Large Language Models}
Recently, there has been an increasing effort to extend the powerful complex reasoning and world knowledge capabilities of LLMs~\cite{chung2022flan-t5, touvron2023llama, zhang2022opt} to other modalities~\cite{chen2023position, zhu2023minigpt4, li2023blip2, instructblip, zhu2023minigpt, chen2023xllm, chen2023minigpt, ye2023mplug}. Among these works, some have aimed to enable LLMs to understand the 3D world.~\cite{xu2023pointllm, han2023imagebind-llm} focus on delving into LLMs' ability to comprehend 3D objects, which can not be directly applied to 3D scenes. 3D-LLM~\cite{hong20243d-llm} is the pioneering work that incorporates the 3D scene into LLM to carry out general 3D understanding tasks. However, by using 3D features constructed through projecting the 2D features of multi-view 2D images extracted by pre-trained 2D Vision-Language Models into 3D space, 3D-LLM struggles to directly capture the complex spatial relationships between objects and the structure of 3D scenes.~\cite{chen2023ll3da, li2023m3dbench} directly extract 3D features from the reconstructed 3D point cloud and support multi-modal visual prompts (text, images, 3D objects) in an MLLM. To alleviate the difficulty LLMs face in understanding complex 3D scenes,~\cite{huang2023chat3d-v2} choose to first explicitly segment the objects in the 3D scene and then perform multi-stage object-aware scene-text alignment to achieve 3D scene understanding. However, due to the lack of large-scale 3D-language alignment data and the intricate content of 3D scenes, although current MLLMs can achieve favorable performance in 3D scene understanding tasks, their localization performance is still significantly behind the 3D localization specialists. Our approach seeks to address this issue by introducing a 3D grounding model to enhance the localization capability of MLLMs.

\section{ScanReason Benchmark}
\label{sec:benchmark}

\subsection{3D Reasoning Grounding Task}
\begin{figure*}[t]
  \centering
   \includegraphics[width=1\linewidth]{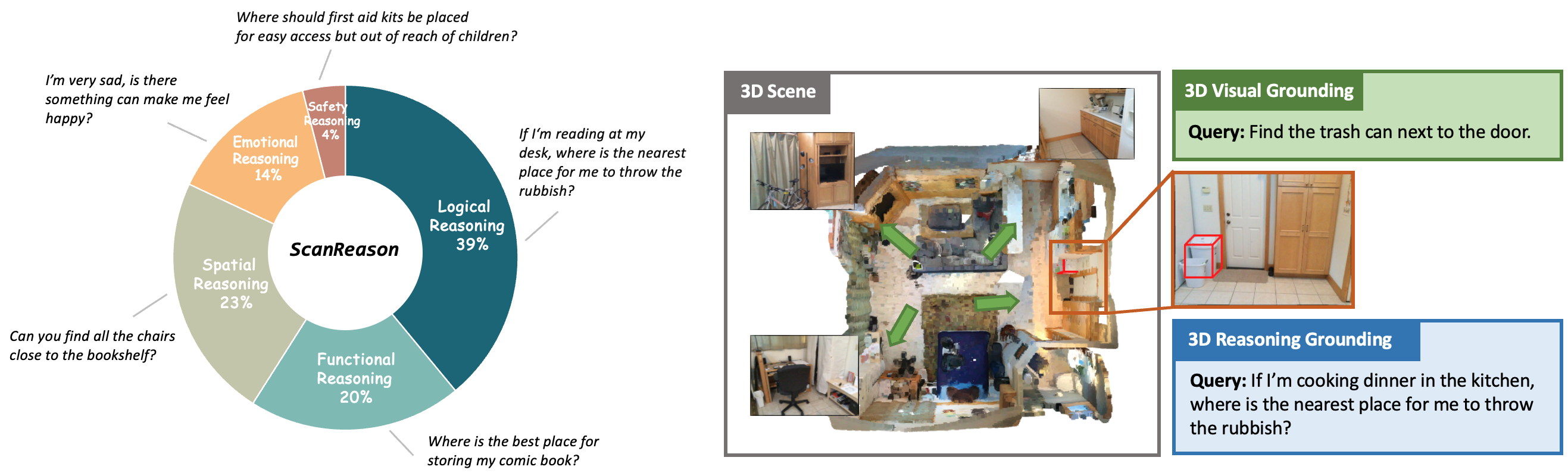}
   \caption{The left side figure shows the overall of our ScanReason dataset. For each reasoning category, we designed different prompts to generate corresponding questions. And the right side figure shows the differences between the traditional 3D visual grounding task and our proposed 3D reasoning grounding task.} 
   \label{fig:task}
\end{figure*}

Given a 3D scene and an implicit free-form query, \taskname{} requires the model to predict the 3D bounding boxes of the target objects as well as the textual answers and explanations. As shown in~\cref{fig:task}, different from traditional 3D visual grounding, the queries of \taskname{} are implicit and complex, requiring strong reasoning, commonsense, and world knowledge. The number of target objects in \taskname{} is flexible and any object satisfying the requirements should be considered as the target object.

\subsection{Question Types}

To comprehensively evaluate the \taskname{} abilities, we define 5 types of questions depending on which type of reasoning is required. \textit{Spatial reasoning} and \textit{functional reasoning} require a fundamental understanding of the 3D physical world, while \textit{logical reasoning}, \textit{emotional reasoning}, and \textit{safety reasoning} are high-level reasoning skills built upon the two fundamental reasoning abilities to address user-centric real-world applications, as shown in~\cref{fig:task}. 

\noindent\textbf{Spatial Reasoning} measures models' understanding of 3D spatial relationships among objects in 3D scene. It encompasses the ability to comprehend the layout of the 3D scene and the 3D location of objects within it, which could serve as the foundation for navigating or planning movements in the 3D environment. 

\noindent\textbf{Functional Reasoning} involves understanding and inferring the purpose, function, or affordance of objects within the 3D scene. For example, functional reasoning allows an embodied agent to recognize that a chair is for sitting, a lamp is for lighting, and a refrigerator is for storing food at low temperatures. Such understanding enables the embodied agent to assist users and to perform complex tasks more effectively (e.g., turning on a lamp when the room gets dark, or navigating to a refrigerator to fetch a drink).

\noindent\textbf{Logical Reasoning} allows an embodied agent to not only understand its environment but also to interact with it in a goal-directed manner. For example, given a question shown in~\cref{fig:task}: ``If I’m cooking dinner in the kitchen, where is the nearest place to throw the rubbish?'', an agent needs to use such reasoning ability to infer the location of objects (in this question, a rubbish bin) based on their function and spatial relationships under the specific setting (the kitchen).

\noindent\textbf{Emotional Reasoning} plays a critical role in human-robot interaction, where the target objects are determined by understanding human emotions, preferences, and behavioral patterns. This ability makes the embodied agents more attuned to the emotional and psychological states of humans, allowing them to provide more personalized, empathetic, and contextually appropriate responses and solutions, such as: ``I'm very sad, is there something that can make me feel happy?'' shown in~\ref{fig:task}.

\noindent\textbf{Safety Reasoning} focuses on preventing harm and ensuring the well-being of humans in the 3D environment. It requires the embodied agent to identify and assess the risk and make safety-aware decisions, such as: ``Where should first aid kits be placed for easy access but out of reach of children?''.

\subsection{Automatic Data Annotation with GPT-4}

We leverage the 3D scenes and bounding box annotations from the EmbodiedScan dataset~\cite{wang2023embodiedscan} and apply GPT-4~\cite{openai2023gpt4} to automatically generate question-answer-location pairs for the five question types respectively. Specifically, we provide GPT-4 with the categories and bounding box locations of all objects in the scene, and ask GPT-4 to generate questions and answers with the target object ids of the provided objects. The details can be found in the Appendix.

In total, our ScanReason dataset consists of 12929 complex reasoning question-answer-location pairs from 1456 scenes, which are split into 11455 training pairs and 1474 validation pairs. All 1474 validation question-answer pairs have been manually verified, including 342 spatial reasoning questions, 287 functional reasoning questions, 581 logical reasoning questions, 211 emotional questions, and 53 safety reasoning questions. We provide detailed statistics and more examples of our dataset in the Appendix.

\subsection{Evaluation Metric}
To evaluate the accuracy of predicted objects and their locations for a flexible number of ground-truth objects, we follow the evaluation metric of 3D object detection tasks. Specifically, we adopt Acc@$k$IoU as our metric, where $k$ is the threshold for the Intersection of Union (IoU) between positive predictions and ground truths. We evaluate the performance under $k=0.25$ in our experiments.

\section{Method}
\label{sec:method}

Solving the task of \taskname{} requires the synergization of the perception, reasoning, and grounding capability of the embodied agent. 
Intuitively, we can first conduct reasoning based on implicit instruction such as ``where should first aid kits be placed?'' and the visual environment. The reasoning process provides us with information about the rough location and semantics of the object we are looking for. Then keeping that information in mind, we look back to the 3D environment to precisely locate the object. For complex scenarios, alternate reasoning and looking back are required for multiple rounds until we obtain the final answer.

\begin{figure*}[t]
  \centering
   \includegraphics[width=1\linewidth]{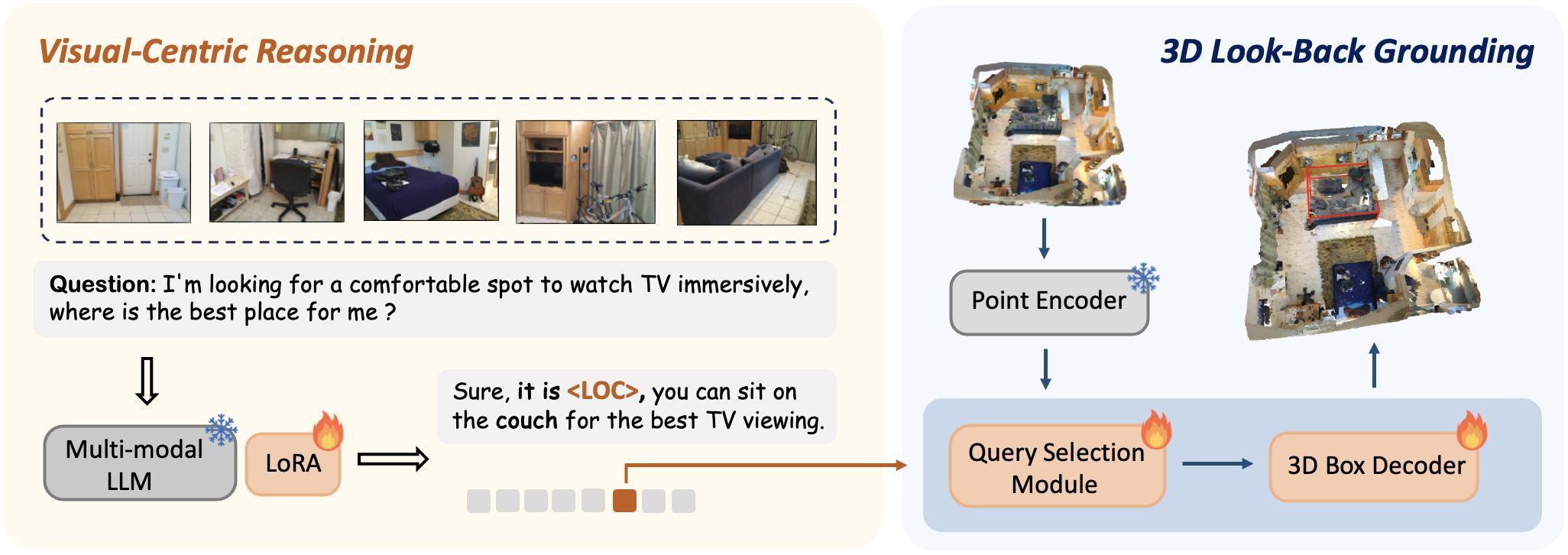}
   \caption{The pipeline of \methodname{}.  Given the 3D scene and human instruction, the visual reasoning module first performs joint 3D scene and instruction reasoning, and then guide the 3D grounding module to look-back the 3D scene and perform target object location. } 
   \label{fig:pipeline}
\end{figure*}

Inspired by this intuition, we propose \methodname{}, consisting of a visual-centric reasoning module and a 3D grounding module with geometry-enhanced look-back, as illustrated in~\cref{fig:pipeline}. 
The visual-centric reasoning module (\cref{subsec:3d_mllm}) performs joint reasoning of language instruction and visual scene, and predicts a special \texttt{<LOC>} token representing the grounding information.
The 3D grounding module (\cref{subsec:3d_grounding_module}) looks back to the original 3D scene with comprehensive geometry information and fine-grained details. It takes the hidden embedding of the \texttt{<LOC>} token containing grounding-related information from the 3D features and eventually predicts the 3D locations of the target objects. 
Furthermore, we propose a Chain-of-Grounding mechanism (CoG) (\cref{subsec:chain_of_grounding_module}), \ie, a chain of interleaved reasoning and grounding steps, to further synergize the grounding and reasoning capability for the \taskname{} task, as illustrated in~\cref{fig:cog}.

\subsection{Visual-Centric Reasoning}\label{subsec:3d_mllm}
Due to the complexity of the 3D scene and user instructions, particularly given the implicit intention behind the human instruction, \methodname{} starts with a visual-centric reasoning module that can perceive the scene, comprehend the human instructions, and conduct joint reasoning of 3D scene and instructions. We believe the reasoning process eventually implies the grounding intention, \ie, implicitly encodes the information indicating the target object to solve the task. Thus, we design the visual-centric reasoning module to predict grounding queries for localizing the target objects in the following 3D grounding module.

Specifically, for simplicity, we leverage 3D-LLM to serve as the visual-centric reasoning module because of its strong reasoning abilities inherited from the LLM. 
Based on the BLIP2 architecture~\cite{li2023blip2}, 3D-LLM uses pre-trained image encoders to extract multi-view 2D image features and back-project them into 3D spaces. The visual features are encoded by the Q-Former to produce 32 tokens as the visual input to the LLM. By leveraging the pre-trained image encoders and the Q-Former, the visual tokens encode rich semantics but lack the 3D structures, spatial interactions, and fine-grained details.
Therefore, instead of directly predicting the object locations by the 3D-LLM, we ask the 3D-LLM to predict the feature representation as the output of the reasoning process, and the predicted feature is further used to ground the target object in the grounding module.

To enable the prediction of the grounding feature, we expand the original vocabulary of the 3D-LLM with a special \texttt{<LOC>} token. The \texttt{<LOC>} token is laden with the contextual scene and the target object information which can guide the 3D grounding module to accurately localize target objects.

\subsection{3D Grounding with Geometry-Enhanced Look-Back}\label{subsec:3d_grounding_module}

Once obtaining the \texttt{<LOC>} token, \methodname{} extracts the last-layer embedding $h_{loc}$ of the \texttt{<LOC>} token and sends it into the 3D grounding module to predict the 3D bounding boxes. 
The 3D grounding module is devised with a ``look-back'' mechanism which allows the model to access the 3D geometry and fine-grained details from a 3D point cloud encoder. The fine-grained geometry-enhanced 3D visual features and $h_{loc}$ are sent into a query selection module to retrieve the most relevant object features. Those features are further decoded into 3D bounding boxes with the 3D box decoder.

\noindent\textbf{3D Visual Encoder}
Unlike the 2D image encoder used for 3D-LLM, the 3D visual encoder directly extracts features from 3D point clouds to capture more geometric and spatial information about the 3D structures and fine-grained details. The powerful 3D visual encoder which captures comprehensive geometry, structure, layout, and fine-grained information is critical to accurate 3D grounding. Subsequently, the 3D features $f_{scene}$ produced by the 3D visual encoder and the grounding feature $h_{loc}$ from 3D-LLM are sent to the query selection module.

\noindent\textbf{Query Selection Module}
We adopt a cross-attention mechanism, where we treat $f_{scene}$ as $Q$ (Query), and $h_{loc}$ as both $K$ (Key) and $V$ (Value), to implicitly obtain a feature-level reasoning activation heatmap. During this scene look-back process, this module roughly locates scene features that have a high response to the \texttt{<LOC>} token. We then select the $k$ most relevant features as the object query $f_{query}$.

\noindent\textbf{3D Box Decoder} is a classical transformer decoder, which consists of M transformer decoder layers, In each decoder layer, the object queries $f_{query}$ go through the text feature $h_{loc}$ cross-attention layer and scene feature $f_{scene}$ cross-attention layer. Finally, the prediction head takes the updated object queries as input and predicts the final 3D locations and matching score.

\noindent\textbf{Discussion} In comparison to previous works~\cite{chen2023shikra, pix2seq, hong20243d-llm, chen2023ll3da} that directly predicts the bounding boxes by the LLM, the extra grounding module has the following advantages:
1) Based on the MLLM reasoning results, it allows the grounding module to perceive the scene again and focus on the region under the implicit guidance of MLLM, adapted to the user queries and the reasoning results.
2) The scene representation perceived in the grounding module can be more precise and fine-grained, which is complementary to the visual-centric reasoning module.
3) The two-step reasoning-grounding pipeline is flexible and can generalized to other types of predicting formats such as segmentation masks (by simply replacing the 3D box decoder with a 3D mask decoder).

\begin{figure*}[t]
  \centering
   \includegraphics[width=1\linewidth]{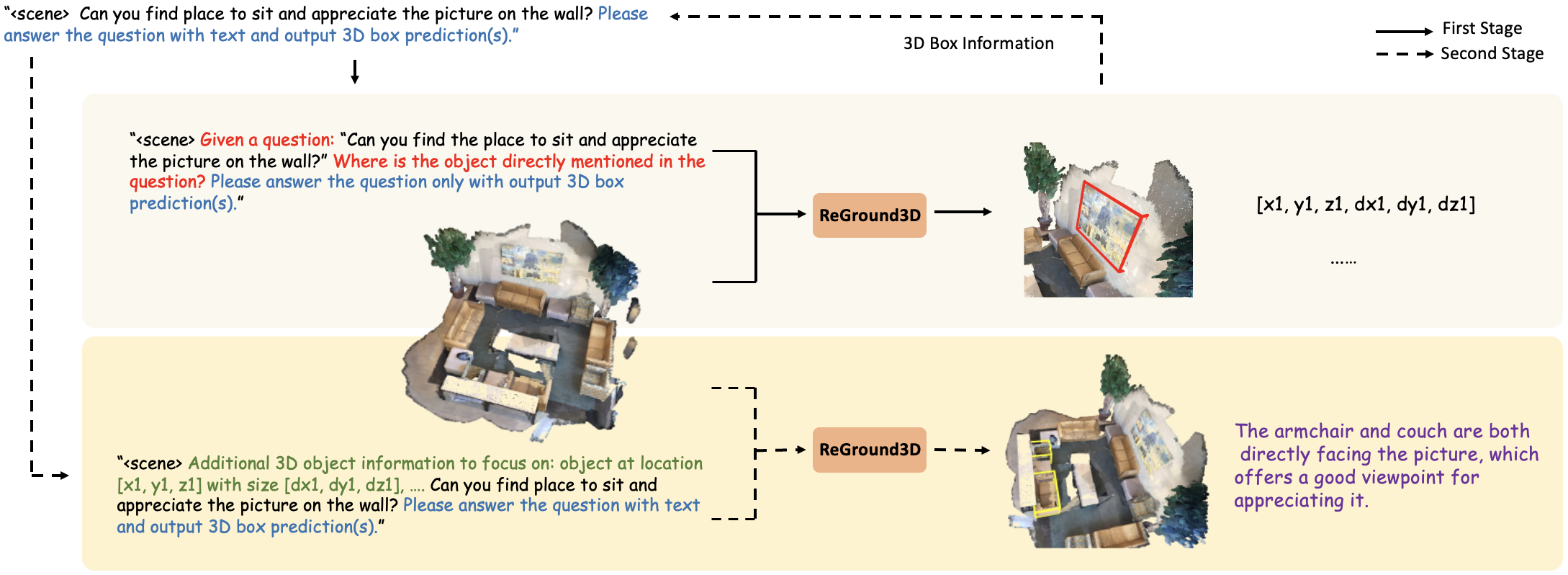}
   \caption{Illustration of Chain-of-Grounding (CoG) Mechanism} 
   \label{fig:cog}
\end{figure*}

\subsection{Chain-of-Grounding Mechanism}\label{subsec:chain_of_grounding_module}

The existing design conducts the reasoning and grounding process sequentially, \ie, the reasoning process is finished before grounding. We argue that the grounding results can also facilitate the reasoning process, especially for those requiring spatial information. Thus, to further synergize the reasoning and grounding process, inspired by chain-of-thought (CoT)~\cite{wei2022chain-of-thought}, we propose \mechanismname{} (CoG), which introduces a chain of interleaved steps of reasoning and grounding to find the targeted objects during inference, as shown in~\cref{fig:cog}. Such a process allows the model to actively find relevant objects that help solve the problem, and then conduct reasoning with the assistance of the additional information of these relevant objects so that the model can more precisely find the target objects.

Specifically, given a question provided by users, CoG translates it into another question of finding the explicitly mentioned objects in the original question. The generated new question is sent into \methodname{} to ground the objects mentioned in the original question in the 3D scene with corresponding confidence scores. An object can be seen as a successfully located object when its confidence score is above the threshold, and the located object information could serve as explicit guidance for 3D-LLM in the next reasoning stage.
As shown in~\cref{fig:cog}, after obtaining the 3D information of objects explicitly mentioned in the original question, the located object information is inserted to update the question. The updated question is then sent to \methodname{} to perform reasoning and grounding to output the target object locations.

\subsection{Instruction Tuning}
\subsubsection{Training Objective}
We use the pre-trained weights of 3D-LLM as the initialization for the visual-centric reasoning module. Except for freezing the 3D visual encoder pre-trained on~\cite{wang2023embodiedscan}, the rest of the parameters in the visual-centric reasoning module and 3D grounding module in our framework are trained in an end-to-end manner. The training supervision is a weighted sum of the next token prediction loss from 3D-LLM and the 3D detection loss from the 3D grounding module. 

\begin{equation}
\mathcal{L} = \lambda_{text}\mathcal{L}_{text} + \lambda_{det}\mathcal{L}_{det}
\end{equation}

\noindent The 3D detection loss is defined following:

\begin{equation}
\mathcal{L}_{det} = \lambda_{IOU}\mathcal{L}_{IOU} \\
+\lambda_{contrast}\mathcal{L}_{contrast}
\end{equation}

\subsubsection{Instruction Tuning Dataset}

We load the pre-trained weights of 3D-LLM and the 3D visual encoder, and finetune the LoRA of 3D-LLM, the query selection module, and the 3D box decoder with an instruction tuning dataset. To construct the instruction tuning dataset, we reformulate the data annotations from existing 3D datasets into question-answer or question-answer-bounding-box pairs. Specifically, the 3D visual grounding data from ScanRefer~\cite{chen2020scanrefer}, SR3D, NR3D~\cite{achlioptas2020referit3d} and the 3D object detection data from EmbodiedScan~\cite{wang2023embodiedscan} are formulated into question-answer-bbox pairs, and the 3D/spatial question answering data from SR3D~\cite{achlioptas2020referit3d}, CLEVR3D~\cite{yan2021clevr3d}, SQA3D~\cite{ma2022sqa3d} are formulated into question-answer pairs without bounding boxes. The information shown in~\cref{tab:instructions} illustrates how we unify the instruction and output with task-specific templates. More details can be found in the Appendix. The reformulated data combined with our proposed ScanReason dataset, serve as the instruction tuning dataset. 

\begin{table*}[t!]
    \caption{We list part of data templates used to train \methodname{} for each task.}
    \label{tab:instructions}
    \centering
    \resizebox{\linewidth}{!}{
    \begin{tabular}{lp{5cm}p{6cm}p{4cm}}
    \toprule[1.2pt]
    \textbf{Task Name} & \textbf{Text Instructions} & \textbf{Output Type Templates} & \textbf{Expected Output} \\ \hline
    \multirow{3}{*}{3D Visual Grounding}    
            & \textcolor[RGB]{215,48,39}{<scene>} Here is a description about an object: "\textcolor[RGB]{13,128,39}{<expr>}", where is the object in the 3D scene? & Please answer the question only with output 3D box prediction(s). & \multirow{3}{*}{It is \textcolor[RGB]{13,108,126}{<LOC>}.}   \\ 
    \hline
    \multirow{6}{*}{3D/Spatial Question Answering}  
            & \textcolor[RGB]{215,48,39}{<scene>} Answer the question: ``\textcolor[RGB]{125,48,39}{<question>}''  &Please answer the question only with text, do not output 3D box prediction(s).         & \multirow{2}{*}{\textcolor[RGB]{125,18,39}{<answer>}.}     \\
            & \textcolor[RGB]{215,48,39}{<scene>} <situation>, \textcolor[RGB]{125,48,39}{<question>}    &Please answer the question only with text, do not output 3D box prediction(s).         & \multirow{2}{*}{\textcolor[RGB]{125,18,39}{<answer>}.}     \\
            & \textcolor[RGB]{215,48,39}{<scene>}\textcolor[RGB]{125,48,39}{<question>}    &Please answer the question only with text, do not output 3D box prediction(s).         & \multirow{2}{*}{\textcolor[RGB]{125,18,39}{<answer>}.}    \\
    \hline
    \multirow{2}{*}{3D Object Detection}      
            & \textcolor[RGB]{215,48,39}{<scene>} Where is the \textcolor[RGB]{69,117,180}{<category>} in this 3D scene?   &Please answer the question only with output 3D box prediction(s). &\multirow{2}{*}{Sure, \textcolor[RGB]{13,108,126}{<LOC>}.}      \\ 
    \hline
    \multirow{2}{*}{3D Reasoning Grounding}
            & \textcolor[RGB]{215,48,39}{<scene>} Answer the question: ``\textcolor[RGB]{125,48,39}{<question>}''    &Please answer the question with text and output 3D box prediction(s).     & \multirow{2}{*}{Sure, \textcolor[RGB]{13,108,126}{<LOC>}, \textcolor[RGB]{13,48,126}{<reason>}}  \\ 
    \bottomrule[1.2pt]
    \end{tabular}
}
\end{table*}

\section{Experiment}
\label{sec:exp}

\subsection{Implementation Details}

\subsubsection{Network Architecture}

For the 3D grounding module, we adopt the pre-trained point cloud encoder as the 3D visual encoder. During the training stage, We use LoRA to efficiently finetune the 3D-LLM to preserve the original 3D scene understanding capability and reduce the computation costs. The number of object queries $k$ in the query selection module is set to 256.

\subsubsection{Training Parameters}
The training is done on 8 NVIDIA A100 GPUs. We adopt the AdamW optimizer with a learning rate of 3e-4 and use a learning rate scheduler WarmupDecayLR with the warmup steps of 100. The total batch size is set to be 16. The loss weight parameters $\lambda_{text}$ and $\lambda_{det}$ in total loss $\mathcal{L}$ are set to 1.0 and 1.0, respectively, and the weight $\lambda_{IOU}$ and $\lambda_{contrast}$ in $\mathcal{L}_{det}$ are set to 1.0 and 1.0.

\subsection{Results Comparison}

\subsubsection{Evaluation on 3D Visual Grounding}

In order to verify the superiority of our model in grounding ability and facilitate comparison between current models, we report the explicit grounding performance on the existing 3D visual grounding task. Since the evaluation settings of Nr3D and Sr3D~\cite{achlioptas2020referit3d} are based on ground-truth object proposals, while ScanRefer~\cite{chen2020scanrefer} requires models to output 3D bounding boxes, we choose ScanRefer as our benchmark for comparison. We divide the existing methods into two categories, one is the grounding model designed specifically for the 3D visual grounding task, and the other is generalist MLLMs which can understand a variety of 3D vision-language tasks. The original 3D-LLM embeds 3D locations in the vocabularies and represents the grounded 3D bounding boxes by a sequence of discrete location tokens. However, since the fine-tuned model of 3D-LLM on ScanRefer and related location tokens implementation are not accessible, we adapt 3D-LLM to directly output 3D numerical coordinates representing 3D bounding boxes by fine-tuning the pre-trained model on our reformulated 3D visual grounding data, denotes as 3D-LLM (vg). As shown in~\cref{tab:3d_vg}, current generalist MLLM models still lag behind the specialist models in terms of grounding ability. By incorporating the 3D grounding module into MLLM, \methodname{} shows the SOTA performance on the traditional 3D visual grounding task. 

\begin{table*}
  \caption{Results on 3D visual grounding task among \methodname{} (ours) and existing methods.
  }
   \label{tab:3d_vg}
    \centering
    \resizebox{0.60\textwidth}{!}{
    \begin{tabular}{c|c|cc}
    \toprule
    \textbf{Type} &\textbf{Methods}	&\textbf{Acc@0.25}	&\textbf{Acc@0.5} \\
    \midrule
    \multirow{6}{*}{\tabincell{c}{Specialists}}     
    &ScanRefer~\cite{chen2020scanrefer}	&37.3	&24.3	\\
    &MVT~\cite{huang2022mvt}	&40.8	&33.3 \\
    &3DVG-Trans~\cite{zhao20213dvg}	&45.9	&34.5	\\
    &ViL3DRel~\cite{chen2022vil3dref}	&47.9	&37.7 \\  
    &BUTD-DETR~\cite{jain2022bottom}	&52.2 &39.8 \\  
    &L3Det~\cite{zhu2023object2scene}	&52.8 &40.2 \\  
    \midrule
    \multirow{5}{*}{\tabincell{c}{Generalized MLLMs}}     
    &LLM-Grounder~\cite{yang2023llm-grounder}	&17.1	&5.3	\\
    &3D-LLM~\cite{hong20243d-llm}   	&30.3	&-   \\
    &3D-LLM(vg)~\cite{hong20243d-llm}	&33.1	&28.7	\\
    &Chat3D-v2~\cite{huang2023chat3d-v2}	&35.9	&30.4 \\
    \midrule  
    ours &\methodname{}	& \textbf{53.1}	&\textbf{41.1}   \\
    \bottomrule
    \end{tabular}}
\end{table*}

\subsubsection{Evaluation on 3D Reasoning Grounding}

\begin{figure*}[t]
  \centering
   \includegraphics[width=1\linewidth]{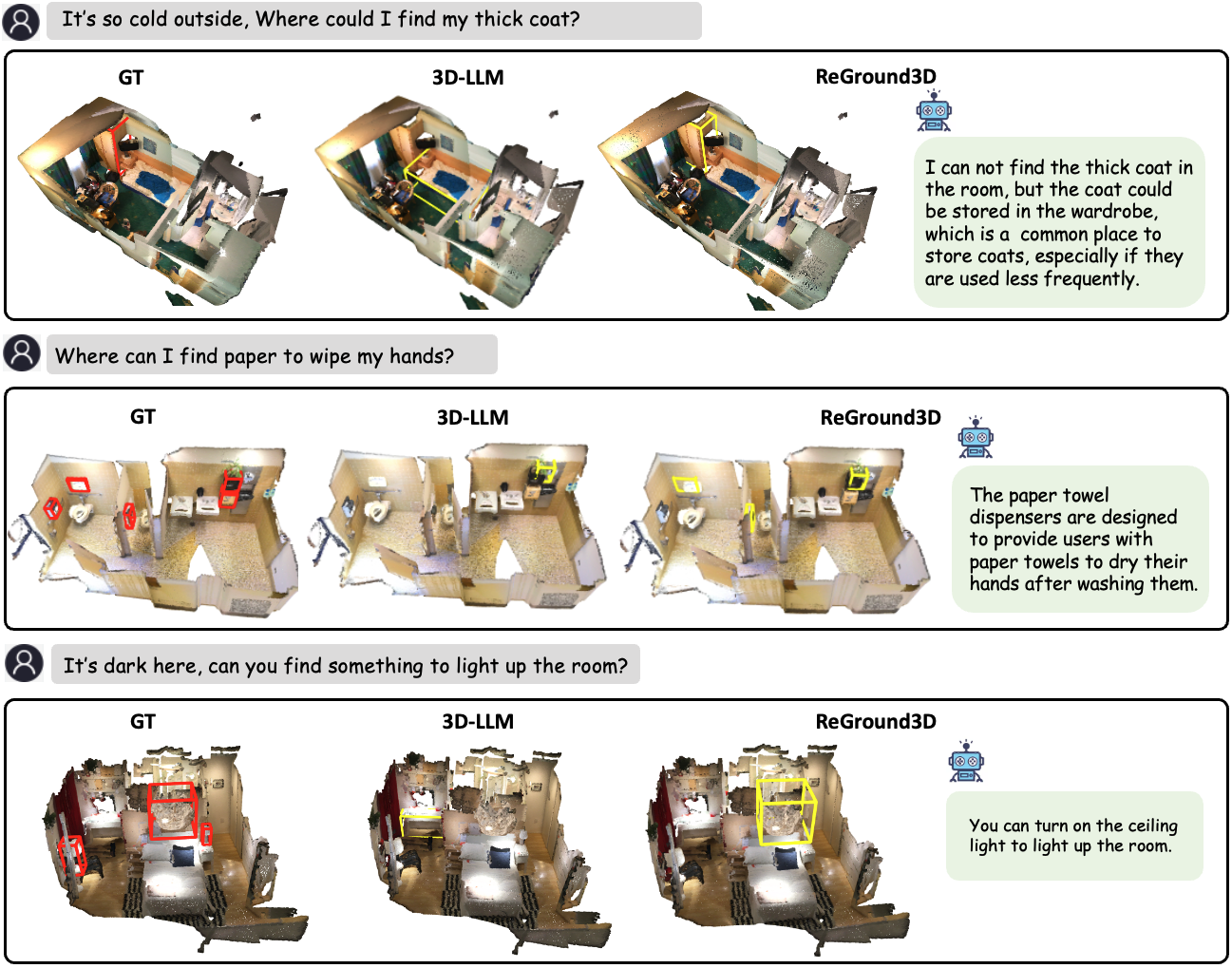}
   \caption{Visualization comparison of \taskname{} capability between \methodname{} and 3D-LLM. Our method could achieve much more accurate grounding results which satisfy the implicit question intention, and give the corresponding explanation at the same time. More illustrations are given in the Appendix. } 
   \label{fig:qualitative}
\end{figure*}

The various visual grounding models rely on explicit text-object alignment in the input object expression to achieve localization, which falls to be applied to \taskname{} task. We performed a comparison between our proposed method \methodname{} and existing MLLM methods, including 3D-LLM(vg) and Chat3D-v2~\cite{huang2023chat3d-v2}. Besides, inspired by Chat-3D v2, which first segments objects, then equips them with unique object identifiers to conduct effective object grounding, we set up a LLM-based 3D reasoning grounding baseline: We first segment the objects from the scene using a 3D instance segmentor~\cite{schult2023mask3d}, then convert the segmented object information including their categories and 3D bounding boxes into text as input of LLM (InternLM2-7B~\cite{team2023internlm}). Besides, to better validate the performance of our model and ensure a fair comparison, we remove the ScanReason dataset from the training data, denoted as \methodname{}*. As shown in~\cref{tab:3lr}, the LLM-based reasoning method (Mask3D~\cite{schult2023mask3d} + InternLM2-7B~\cite{team2023internlm}) possesses a very strong functional reasoning ability, but struggles to understand of 3D spatial relationship. Additionally, irrespective of whether ScanReason is used in training, our model significantly outperforms the existing MLLMs. By synergizing the reasoning and grounding process utilizing the Chain-of-Grounding (CoG) mechanism, the \taskname{} performance of \methodname{} can be further improved (from 28.98 to 30.62), especially on the spatial reasoning and logical reasoning questions. The qualitative results comparison shown in~\cref{fig:qualitative} demonstrates the superiority of our method in reasoning human complex instruction based on the 3D scene.

\begin{table*}[t!]
  \caption{Results (Acc@0.25) on \taskname{} task among \methodname{} (ours) and existing methods.
  }
   \label{tab:3lr}
    \centering
    \resizebox{1.0\textwidth}{!}{
    \begin{tabular}{l|l|ccccc|c}
    \toprule
    \textbf{Methods}	& \textbf{LLM} &\textbf{Spatial}	&\textbf{Functional } &\textbf{Logical } &\textbf{Emotional} &\textbf{Safety} &\textbf{Overall} \\
    \midrule
    Mask3D~\cite{schult2023mask3d} + InternLM2-7B~\cite{team2023internlm}  &InternLM2-7B &10.34 &36.12 &9.98  &8.21 &8.99	&14.86\\
    \midrule
    3D-LLM(vg)~\cite{hong20243d-llm}  &FlanT5$_\text{XL}$-3B &18.31	&17.42 	&10.97  &8.12 &6.33	&13.29\\
    Chat3D-v2~\cite{huang2023chat3d-v2} &Vicuna-7B &20.21	&18.39 &11.32  &7.98  &9.88 &14.98\\
    \midrule  
    \methodname{}*	&FlanT5$_\text{XL}$-3B & 30.76	&29.8  &18.67  &19.22 &17.12  &23.27\\
    \methodname{}	&FlanT5$_\text{XL}$-3B & 32.98 & 36.23
    &26.99  &23.12  &22.98  &28.98 \\
    \methodname{} (CoG) &FlanT5$_\text{XL}$-3B & \textbf{34.71}	&\textbf{36.79} 
    &\textbf{29.11}  &\textbf{24.03}  &\textbf{23.21}  &\textbf{30.62}\\
    \bottomrule
    \end{tabular}}
\end{table*}

\subsection{Ablation Study}

In this section, we conduct an extensive ablation study to verify the effectiveness of each component.

\subsubsection{Effectiveness of 3D Grounding Module}

In order to more comprehensively verify the effectiveness of the 3D grounding module we proposed, we step by step verify the performance changes from 3D-LLM to \methodname{}. Apart from the 3D-LLM(vg), we fine-tuned the 3D-LLM model respectively on full reformulated existing data and all instruction tuning data, including ScanReason, denoted as 3D-LLM (full) and 3D-LLM (full+sr). All the \texttt{<LOC>} tokens in answers of the data used for fine-tuning 3D-LLM have been converted to the numerical coordinates of the corresponding boxes. The results in~\cref{tab:3d_ground_module} show that with the same tuning dataset, our \methodname{} achieves far superior performance to 3D-LLM (28.98 vs. 19.21) by introducing the 3D grounding module.

\subsubsection{Effectiveness of Instruction Tuning Dataset}

\cref{tab:traindata} showcases the impact of different training data types on 3D visual grounding performance. 3D Object Detection (3D OD) provides the explicit 3D semantic category and visual alignment whereas 3D Question Answering (3D QA) data injects the basic 3D scene understanding ability into the model, which has certain benefits for the visual grounding ability of the model. In addition, we find that training with \taskname{} dataset can further improve the performance on 3D visual grounding.

\begin{table}[t!]
    \centering   
    \caption{Ablation study on effectiveness of 3D grounding module and training data}
    \begin{subtable}{0.45\linewidth}
        \centering
        \caption{When using the same tuning dataset, \methodname{} achieves far better performance on ScanReason than 3D-LLM.}
        \resizebox{!}{12mm}{
            \begin{tabular}{l|c}
            \toprule
            \textbf{Methods}	&\textbf{Acc@0.25} \\
            \midrule
            3D-LLM(vg)~\cite{hong20243d-llm}             &13.29		\\
            3D-LLM(full)~\cite{hong20243d-llm}            &15.31		\\
            3D-LLM(full+sr)~\cite{hong20243d-llm}         &19.21	\\
            \methodname{}(full+sr)	& \textbf{28.98}    \\
            \bottomrule
            \end{tabular}
        }
        \label{tab:3d_ground_module}
    \end{subtable}
    \hfill
    \begin{subtable}{0.52\linewidth}
        \centering
        
        \caption{Ablation study on training data. We evaluate through the metric of accuracy on the val set of the ScanRefer dataset.}
        \resizebox{!}{13mm}{
            \begin{tabular}{cccccc}
            \toprule
            \multicolumn{4}{c}{\textbf{Dataset}} & \multirow{2}{*}{\textbf{Acc@0.25}} & \multirow{2}{*}{\textbf{Acc@0.5}} \\
                                   \textbf{VG}      & \textbf{OD}    & \textbf{VQA}  & \textbf{RD}        \\ \midrule
                &\Checkmark        & \Checkmark       &                    & 19.3  & 14.2 \\
             \Checkmark &                           & \Checkmark  &  & 48.7  & 37.6  \\
             \Checkmark & \Checkmark &                        &   & 49.2   & 38.1    \\
             \Checkmark & \Checkmark & \Checkmark              &   & 51.8   & 39.4    \\
            \Checkmark & \Checkmark & \Checkmark & \Checkmark & \textbf{53.1}  &\textbf{41.1}    \\ \bottomrule
            \end{tabular}
        }
        \label{tab:traindata}
    \end{subtable}
    \label{tab:combined}
\end{table}

\subsubsection{Discussion of CoG Mechanism}

While the CoG Mechanism boosts the performance with interleaved reasoning and grounding steps during inference, it uses the relevant object information explicitly presented in the question to help find the target objects. A natural question arises: if we input the information of all the objects instead of relevant objects into \methodname{} during CoG, will this make the model more accurately find the target objects? We first use the existing 3D Segmentor~\cite{schult2023mask3d} to segment all objects in the scene, then update the original question using all the object 3D bounding boxes information according to the template in~\cref{subsec:chain_of_grounding_module} and send into \methodname{}. However, experiments show that using all the object information will instead reduce the \taskname{} performance from 28.98 to 27.67. One possible reason could be the model's attention is dispersed by too many irrelevant objects.

\section{Conclusion}
\label{sec:conclusion}

This paper introduces a new 3D vision language learning task: \taskname{}, which requires the model to perform active reasoning over complex and implicit human instruction, localize the target objects, and give corresponding explanations. Furthermore, we propose ScanReason, a new dataset and benchmark to further unlock and thoroughly evaluate the \taskname{} capability. Based on this dataset, we propose a novel approach: \methodname{}, which utilizes the strong reasoning capability of MLLM guiding the 3D grounding module to obtain accurate object locations, and a Chain of Grounding (CoG) mechanism is presented to further boost the performance with interleaved reasoning and grounding steps during inference. We believe that our work will further the natural interaction between embodied agents and humans in open 3D environments. For the current ScanReason benchmark, we find that the questions in three high-level 3D reasoning categories may have overlaps. For a certain reasoning question, similar questions may appear in one or two other categories. We leave the problem as a future challenge for better reasoning grounding ability evaluation.


%
%
\newpage

\section*{Acknowledgements}
This work is supported in part by HKU Startup Fund, HKU Seed Fund for Basic Research, HKU Seed Fund for Translational and Applied Research, HKU IDS research Seed Fund, and HKU Fintech Academy R\&D Funding.

\bibliographystyle{splncs04}
\bibliography{main}

\end{document}